\documentclass{article}

\PassOptionsToPackage{numbers, compress}{natbib}
\usepackage[utf8]{inputenc}
\usepackage[T1]{fontenc}
\usepackage{lmodern}

\usepackage{graphicx}
\usepackage{subcaption}
\usepackage[preprint]{neurips_2024}
\usepackage{algorithm}
\usepackage{algorithmic}
\usepackage[numbers]{natbib}

\usepackage[utf8]{inputenc} 
\usepackage[T1]{fontenc}    
\usepackage{hyperref}       
\usepackage{url}            
\usepackage{booktabs}       
\usepackage{amsfonts}       
\usepackage{nicefrac}       
\usepackage{microtype}      
\usepackage{xcolor}         
\usepackage{booktabs}
\usepackage{tabularx}
\usepackage{amsmath}
\usepackage{float}

\title{AmCLR: Unified Augmented Learning for Cross-Modal Representations}

\author{%
  Ajay Jagannath\textsuperscript{†,a}\thanks{Order of writing names of authors decided randomly.}, 
  Aayush Upadhyay\textsuperscript{†,a}, 
  Anant Mehta\textsuperscript{†,a}\\
  \textsuperscript{†}First authorship is shared equally among all authors.\\
  \textsuperscript{a}Department of Computer Science \& Engineering, Texas A\&M University, College Station, USA\\
  \texttt{ajayjagan2511@tamu.edu, aaupadhy@tamu.edu, anant\_mehta@tamu.edu}
}

\begin{document}

\maketitle

\begin{abstract}
Contrastive learning has emerged as a pivotal framework for representation learning, underpinning advances in both unimodal and bimodal applications like SimCLR and CLIP. To address fundamental limitations like large batch size dependency and bimodality, methods such as SogCLR leverage stochastic optimization for the global contrastive objective. Inspired by SogCLR's efficiency and adaptability, we introduce \textbf{AmCLR} and \textbf{xAmCLR}—objectives tailored for bimodal vision-language models to further enhance the robustness of contrastive learning. AmCLR integrates diverse augmentations, including text paraphrasing and image transformations, to reinforce the alignment of contrastive representations, keeping batch size limited to a few hundred samples unlike CLIP which needs batch size of 32,768 to produce reasonable results. xAmCLR further extends this paradigm by incorporating intra-modal alignments between original and augmented modalities for richer feature learning. These advancements yield a more resilient and generalizable contrastive learning process, aimed at overcoming bottlenecks in scaling and augmentative diversity. Since we have built our framework on the existing SogCLR, we are able to demonstrate improved representation quality with fewer computational resources, establishing a foundation for scalable and robust multi-modal learning. For more details, please visit: \href{https://github.com/aaupadhy/AmCLR/tree/project}{https://github.com/AmCLR}.
\end{abstract}

\section{Introduction}
Self-supervised learning (SSL) has revolutionized representation learning by eliminating the need for labeled data and leveraging intrinsic data structures for pretraining deep neural networks. While SSL has shown immense success in unimodal settings \cite{chen2020simple} such as computer vision and natural language processing, its application in bimodal settings, where alignment between two distinct modalities (e.g., vision and language) is critical, has gained significant attention in recent years. The bimodal SSL paradigm is particularly impactful in applications like image-caption retrieval, visual question answering, and cross-modal content understanding.

One of the most prominent frameworks for bimodal SSL is CLIP \cite{radford2021learning}, which uses a contrastive learning objective to align image and text representations. CLIP trains vision and language encoders simultaneously by maximizing the similarity between paired image-text representations (positive pairs) and minimizing the similarity with unpaired combinations (negative pairs). This simple yet effective approach allows CLIP to achieve state-of-the-art performance in zero-shot learning and retrieval tasks on popular benchmarks like ImageNet \cite{deng2009imagenet} and MS COCO \cite{lin2014microsoft}. However, despite its success, CLIP's reliance on extremely large batch sizes (e.g., 32,768) poses practical challenges, including high memory requirements and computational overhead.

To address the limitations of large batch sizes, techniques like SogCLR by yuan $et$ $al.$ have been developed in the unimodal as well as bimodal setting \cite{yuan2022provable}. SogCLR introduces a memory-efficient stochastic optimization algorithm for contrastive learning that eliminates the dependency on large batches by optimizing a global contrastive objective. While SogCLR has primarily focused on vision-based SSL, its principles can be extended to bimodal settings to address challenges in aligning image and text representations. iSogCLR by Qiu $et$ $al.$ used distributionally robust objective to calculate individual temperatures for each sample, there exploring a new paradigm to approach the minimization of contrastive loss \cite{qiu2023not}.

The core of contrastive learning lies in its loss function \cite{mehta2023benchmarking}, which operates by attracting positive pairs representations of corresponding inputs (e.g., an image and its associated caption) while repelling negative pairs formed by unrelated combinations. This mechanism ensures that representations from different modalities are well-aligned for paired inputs and distinct for unpaired ones, thereby enabling robust cross-modal understanding.

In this work, we focus exclusively on bimodal contrastive learning, building on the foundational principles of CLIP and SogCLR. We aim to enhance the robustness and generalizability of bimodal alignment through improved contrastive objectives and augmentation strategies. Our contributions are centered on addressing the computational challenges and augmentative diversity required for scalable and effective multimodal learning. \\

Some key remarks about our work:
\begin{itemize}
    \item To test our proposed objectives with limited computational resources, we use a 100k subset of the Conceptual Captions 3M \cite{sharma2018conceptual} (CC3M) dataset for training. For validation, we evaluate on the MSCOCO validation dataset for retrieval tasks and the ImageNet validation dataset for zero-shot classification tasks. 

    \item The model configuration includes a ResNet-50 \cite{he2016deep} pretrained on ImageNet as the image encoder and a DistilBERT \cite{sanh2019distilbert} pretrained on BookCorpus and English Wikipedia as the text encoder. Even after using pretrained encoders with no additional pretraining and fewer trainable parameters, we show competitive performance \& efficiency, thus saving computate cost.

    \item AmCLR and xAmCLR, when paired with AdamW \cite{loshchilov2017decoupled} and AdamP \cite{heo2020adamp}, outperform SogCLR and iSogCLR across all tasks. In Retrieval (Text), AmCLR with AdamW achieves a Top-1 accuracy of 14.64\%, surpassing SogCLR (13.1\%) by 1.54\%, while xAmCLR with AdamW shows a 1.04\% improvement with 14.14\%. Similarly, AmCLR with AdamP achieves a Top-1 accuracy of 14.54\%, which is 2.54\% higher than SogCLR's 12\% and xAmCLR with AdamP achieves a Top-1 accuracy of 13.62\%, showing a 1.62\% improvement over SogCLR. For Retrieval (Image), AmCLR with AdamW (11.08\%) outperforms SogCLR (10.06\%) by 1.02\%, and xAmCLR with AdamW (11.14\%) improves by 1.08\%, while AmCLR with AdamP achieves a Top-1 accuracy of 11.46\%, surpassing SogCLR's 9.32\% by 2.14\%, and xAmCLR with AdamP achieves a Top-1 accuracy of 10.43\%, which is 1.11\% higher than SogCLR.

    \item In Zero-shot classification, AmCLR achieves a Top-1 accuracy of 25.87\%, a 1.59\% increase over SogCLR (24.28\%) with AdamW and a Top-1 accuracy of 25.08\%, a 2.86\% increase over SogCLR (22.22\%) with AdamP, while xAmCLR shows a 1.33\% improvement with 25.61\% with AdamW and 3.6\% improvement with 25.82\% with AdamP.
\end{itemize}

To the best of our knowledge, this is the first work to propose a bimodal augmentation approach aimed at enhancing the accuracy of SogCLR. We hope that this paper will inspire future research efforts focused on developing improved algorithms for optimizing the global contrastive objective.

\section{Related Works}
Self-supervised learning (SSL) has become a cornerstone for representation learning in bimodal settings, particularly in vision-and-language pretraining (VLP) \cite{gan2022vision} . These approaches utilize paired image-text data to train models that effectively align the modalities, enabling applications such as image-caption retrieval and zero-shot classification. At the heart of these methods is contrastive learning, which aligns embeddings of positive image-text pairs while pushing apart those of unpaired combinations. This section reviews key advancements in bimodal contrastive learning.

The foundational CLIP model introduced a contrastive learning framework for jointly training image and text encoders. Given a mini-batch of $m$ image-text pairs $B = \{(x_1, z_1), \dots, (x_m, z_m)\}$, the contrastive loss contrasts positive pairs with all negatives in the batch. For an image-text pair $(x_i, z_i)$, the loss is defined as:

\begin{equation}
\ell_1(x_i, z_j; \tau) = \exp\left( \frac{h_i(w)^\top e_j(w) - h_i(w)^\top e_i(w)}{\tau} \right)
\end{equation}
and
\begin{equation}
\ell_2(z_i, x_j; \tau) = \exp\left( \frac{e_i(w)^\top h_j(w) - e_i(w)^\top h_i(w)}{\tau} \right),
\end{equation}
where $h_i(w)$ and $e_i(w)$ represent the embeddings of the image $x_i$ and text $z_i$, respectively, $\tau$ is the temperature parameter, and $w$ are the model parameters.

The batch contrastive loss for the pair $(x_i, z_i)$ is then:

\begin{equation}
L(w, \tau, x_i, z_i, B) = \log \left( \sum_{z_j \in B} \ell_1(x_i, z_j; \tau) \right) + 
\log \left( \sum_{x_j \in B} \ell_2(z_i, x_j; \tau) \right).
\end{equation}

Finally, the loss is averaged across the mini-batch:

\begin{equation}
L(w, \tau, B) = \frac{1}{m} \sum_{(x_i, z_i) \in B} L(w, \tau, x_i, z_i, B).
\end{equation}

The optimization involves jointly learning the model parameters $w$ \cite{mehta2023multi} and the temperature parameter $\tau$, ensuring robust alignment between the image and text modalities.

Building on CLIP proposed by Radford $et$ $al.$, CyCLIP \cite{goel2022cyclip} by Goel $et$ $al.$ introduced cyclic consistency to enhance cross-modal alignment. This ensures that semantic relations in one modality are preserved when mapped into the other modality's embedding space. Such fine-grained cross-modal interactions improve downstream task performance by addressing semantic mismatches inherent in web-crawled datasets.

Other works such as ALIGN \cite{zhang2022align}(Han $et$ $al.$) and Smeu $et$ $al.$'s DeCLIP \cite{smeu2024declip} extended CLIP by utilizing larger datasets or additional supervision signals to further enhance performance. SLIP \cite{mu2022slip} and FILIP \cite{yao2021filip} introduced modifications to the training pipeline to improve cross-modal feature extraction. These models emphasize either fine-grained semantic alignment or augment the representation space to better capture complex relationships between modalities.

Contrastive loss functions remain central to these methods. A common formulation, the InfoNCE \cite{rusak2024infonce} loss, can be written as:
\[
L = -\log \frac{\exp(\text{sim}(x_i, z_i)/\tau)}{\sum_{j=1}^m \exp(\text{sim}(x_i, z_j)/\tau)},
\]
where $\text{sim}(\cdot, \cdot)$ is typically the cosine similarity. The loss encourages alignment of positive pairs $(x_i, z_i)$ while pushing apart negatives $(x_i, z_j)$ for $i \neq j$. Extensions of this loss, as employed in models like CLIP and CyCLIP, ensure effective representation learning in bimodal settings.

While these methods have achieved remarkable performance, they often rely on large-scale datasets and computational resources. Recent efforts like SogCLR and iSogCLR have addressed the challenges of optimizing contrastive losses with smaller batch sizes and memory constraints. Although originally designed for unimodal tasks, their principles have shown promise in bimodal scenarios, enabling efficient and scalable learning for vision-and-language tasks.

In next section we discuss about global contrastive loss used by SogCLR and built on our frameworks. iSogCLR is just using individual temperature parameter for each sample. We plan to further extend our approach with iSogCLR but we don't discuss it in this study.

\section{Proposed Framework for Optimizing Global Bimodal Contrastive Loss}
In this work, we introduce two algorithms built on the SogCLR framework \cite{yuan2022provable}. Before discussing our proposed algorithms, it is essential to first address the foundational SogCLR loss function. SogCLR addresses the challenges of optimizing the global contrastive loss by offering a memory-efficient stochastic optimization algorithm. This method enables efficient handling of large datasets without relying on large mini-batches. We chose to build upon the SogCLR loss because of these advantages. Understanding the SogCLR loss is critical, as it forms the basis for our proposed models, which aim to improve upon SogCLR's performance while maintaining its efficiency. 

The SogCLR loss function is formulated to optimize the global contrastive loss over the entire dataset, avoiding the need for large mini-batches. The objective function can be defined as:

\begin{equation}
F(w) = -\frac{\tau}{n} \sum_{i=1}^{n} \log \frac{\exp \left(E_I(x_i)^\top E_T(t_i) / \tau \right)}{\sum_{t \in D} \exp \left(E_I(x_i)^\top E_T(t) / \tau \right)} - \frac{\tau}{n} \sum_{i=1}^{n} \log \frac{\exp \left(E_I(x_i)^\top E_T(t_i) / \tau \right)}{\sum_{x \in D} \exp \left(E_I(x)^\top E_T(t_i) / \tau \right)},
\end{equation}

where $E_I(x_i)$ and $E_T(t_i)$ are the image and text embeddings for the sample $(x_i, t_i)$, and $D$ is the entire dataset. $\tau$ is temperature. Here, $n$ refers to the size of full dataset.

Given the large size of the dataset $D$, the challenge lies in computing the terms:

\[
g(w; x_i) = \mathbb{E}_{t \sim D} \left[ \exp \left( E_I(x_i)^\top E_T(t) / \tau \right) \right]
\]
and
\[
g(w; t_i) = \mathbb{E}_{x \sim D} \left[ \exp \left( E_I(x)^\top E_T(t_i) / \tau \right) \right].
\]

To address this, we propose a stochastic gradient estimator to handle the large dataset efficiently. The estimator is given by:

\begin{equation}
m_t = - \frac{1}{B} \sum_{i \in B} E_I(x_i)^\top E_T(t_i) + \frac{1}{B} \sum_{i \in B} \tau/ u_{I_{i,t}} \nabla g(w_t; x_i, B) + \tau/ u_{T_{i,t}} \nabla g(w_t; t_i, B),
\end{equation}

where $g(w; x_i, B)$ and $g(w; t_i, B)$ are the mini-batch estimators of $g(w; x_i)$ and $g(w; t_i)$, respectively. The scalars $u_{I_{i,t}}$ and $u_{T_{i,t}}$ are updated for the sampled data according to:

\begin{equation}
u_{I_{i,t+1}} = (1 - \gamma) u_{I_{i,t}} + \gamma g(w_t; x_i, B),
\end{equation}
\begin{equation}
u_{T_{i,t+1}} = (1 - \gamma) u_{T_{i,t}} + \gamma g(w_t; t_i, B),
\end{equation}

where $\gamma$ is a hyperparameter controlling the update rate. This formulation enables the use of the full dataset for analysis, rather than relying on mini-batches, improving both the efficiency and robustness of the training process.

Finally, the model parameters $w$ are updated using an Adam-style or momentum-style update.

This formulation of SogCLR forms the foundation of our proposed algorithms. By building on SogCLR, we aim to further improve its performance and generalizability, particularly in the context of large-scale datasets with long-tailed distributions, which is a key challenge in vision-and-language pretraining. Our proposed enhancements retain the memory efficiency and scalability of SogCLR while offering improved performance in handling multimodal data.

\subsection{AmCLR}
In AmCLR, we introduce augmentations by randomly selecting $\omega$ image augmentations from a set of $P1$ augmentations for each image. Let us consider the space of all possible text paraphrases for a given text as $P2$. So, now $\omega$ augmented texts or paraphrasings are sampled from a set of $P2$ augmentations for each text. From these, we create combinations of image and text pairs for various contrastive losses. The intuition behind this approach is that by generating additional augmented versions of each image-text pair in a batch, the model can learn more robust and generalized representations. This is particularly important because small batch sizes can limit the diversity of data that the model is exposed to during training. By augmenting each data point, we effectively increase the variety of training examples without needing to increase the actual batch size. Thus, we augment each image and its corresponding text across the dataset in a batch-wise manner. 

One thing to note is that $\omega$ should be much smaller than the batch size, as excessive augmentations could create as many image-text pairs per batch as in the full dataset, nullifying SogCLR’s efficient small-batch approximation of the global contrastive loss.

\[
\text{Let } \omega \text{ be the number of augmentations per modality.}
\]
\[
\text{So, each modality (image or text) has } (\omega + 1) \text{ variations (original + augmentations).}
\]
\[
\text{Total combinations for both image-to-text and text-to-image are:}
\]
\[
\text{} \kappa = 2 \times (\omega + 1) \times (\omega + 1)
\]
where $\kappa$ is the total number of combinations.\\

Now, the new global contrastive loss spread across all batches can be written as follows:
\begin{align}
F_1(w;B) &= -\frac{\tau}{m} \sum_{i=1}^{m} \log \frac{\exp \left( E_I(x_i)^\top E_T(t_i) / \tau \right)}{\sum_{t \neq t_i} \exp \left( E_I(x_i)^\top E_T(t_j) / \tau \right)}, \\
F_2(w;B) &= -\frac{\tau}{m} \sum_{i=1}^{m} \log \frac{\exp \left( E_I(x_i)^\top E_T(\hat{t}_i) / \tau \right)}{\sum_{\hat{t} \neq \hat{t}_i} \exp \left( E_I(x_i)^\top E_T(\hat{t}_j) / \tau \right)}, \\
F_3(w;B) &= -\frac{\tau}{m} \sum_{i=1}^{m} \log \frac{\exp \left( E_I(\hat{x}_i)^\top E_T(t_i) / \tau \right)}{\sum_{t \neq t_i} \exp \left( E_I(\hat{x}_i)^\top E_T(t_j) / \tau \right)}, \\
F_4(w;B) &= -\frac{\tau}{m} \sum_{i=1}^{m} \log \frac{\exp \left( E_I(\hat{x}_i)^\top E_T(\hat{t}_i) / \tau \right)}{\sum_{\hat{t} \neq \hat{t}_i} \exp \left( E_I(\hat{x}_i)^\top E_T(\hat{t}_j) / \tau \right)}, \\
F_5(w;B) &= -\frac{\tau}{m} \sum_{i=1}^{m} \log \frac{\exp \left( E_I(x_i)^\top E_T(t_i) / \tau \right)}{\sum_{x \neq x_i} \exp \left( E_I(x_j)^\top E_T(t_i) / \tau \right)}, \\
\vdots &\notag \\
F_\kappa(w; \beta) &\text{ (extends similarly...)}
\end{align}

Here, \( B = \{(x_i, t_i)\}_{i=1}^m \) denote a batch of size \( m \), where \( x_i \) and \( t_i \) represent the image and text pairs in the batch. Let's consider $n$ as the size of the dataset $D$. For a given batch with \( F(w; B) \) representing batchwise losses  with $\tau$ as temperature hypermeter, the total loss is:

\begin{equation}
F(w; B) = \sum_{k=1}^{\kappa} F_k(w; B),
\end{equation}

where \( F_k(w; B) \) corresponds to the \( k \)-th combination of original and augmented image and text pairs as previously defined. Since some batches, like the original samples, are sampled uniformly from the dataset \( D \), and others are augmented versions of the batched data, the expectation of \( F(w; B) \) over all possible batches serves as a fair estimate of the overall dataset loss \( F(w) \):

\begin{equation}
F(w) = \mathbb{E}_{B \sim D} \left[ F(w; B) \right].
\end{equation}

This holds because the sampling strategy ensures that all data points in \( D \) are equally likely to appear in the batches \( B \). So, the overall dataset loss \( F(w) \) can be expressed as:

\begin{equation}
F(w) = \frac{m}{n} \sum_{B \in \mathcal{D}} F(w; B),
\end{equation}

Now, \( F(w; B) \) can be decomposed as:
\begin{equation}
F(w; B) = g_1(w; B) + g_2(w; B),
\end{equation}

where \( g_1(w; B) \) and \( g_2(w; B) \) are defined as follows:
\[
g_1(w; B) = F_p(w; B) + F_q(w; B),
\]
and
\[
g_2(w; B) = \sum_{k=1, k \neq p, q}^{\kappa} F_k(w; B).
\]

Here, \( g_1(w; B) \) is deterministic since it only depends on \( B \subseteq \mathcal{D} \), with no randomness from additional sampling. We include \( F_p=(w; B) \) and \( F_q=(w; B) \) denote those two combinations which directly use batch image and texts.

However, \( g_2(w; B) \), involves the summation of terms \( F_k(w; B) \), where \( F_k(w; B) \) is a random variable due to its dependence on randomness in augmentation selection. This makes \( g_2(w; B) \) a stochastic component.

Combining these terms, we can rewrite \( F(w) \) as:
\begin{equation}
F(w) = \mathbb{E}_{B \sim \mathcal{D}} \left[ F_p(w; B) + F_q(w; B) \right]
+ \mathbb{E}_{Image \sim \mathcal P1, Text \sim \mathcal P2, B \sim \mathcal{D}} \left[ \sum_{\substack{k=1 \\ k \neq p, q}}^{\kappa} F_k(w; B) \right].
\end{equation}
Here, the first term represents the expectation over the deterministic part of the loss \( F_p(w; D) + F_q(w; D) \). The second term involves a nested expectation due to randomness in data augmentations.

The gradient estimator \( m_t \) for the full dataset across each batch \( F(w) \) can be expressed same as used in SogCLR.

\begin{equation}
m_t = - \frac{1}{|B|} \sum_{i \in B} E_I(x_i)^\top E_T(t_i) + \frac{1}{|B|} \sum_{i \in B} \tau/ u_{I_{i,t}} \nabla g(w_t; x_i, B) + \tau/ u_{T_{i,t}} \nabla g(w_t; t_i, B),
\end{equation}

where $g(w; x_i, B)$ and $g(w; t_i, B)$, as seen in algorithm \ref{algo:AmCLR}, are accumulated gradients for all the $\kappa$ cases per batch and the batch estimators of $g(w; x_i)$ and $g(w; t_i)$, respectively. The scalars $u_{I_{i,t}}$ and $u_{T_{i,t}}$ are updated for the sampled data according to:
\begin{equation}
u_{I_{i,t+1}} = (1 - \gamma) u_{I_{i,t}} + \gamma g(w_t; x_i, B),
\end{equation}
\begin{equation}
u_{T_{i,t+1}} = (1 - \gamma) u_{T_{i,t}} + \gamma g(w_t; t_i, B),
\end{equation}
Finally, the model parameters are updated using an AdamW-style update rule:
\begin{equation}
w_{t+1} = w_t - \eta v_t,
\end{equation}

where $v_t$ is the momentum term and $\eta$ is the learning rate.

\begin{algorithm}[H]
\caption{AmCLR}
\label{algo:AmCLR}
\begin{algorithmic}[1]
\REQUIRE $\mathbf{w}_0 \in \mathbb{R}^d$, Initialize $\mathbf{u}_0^I, \mathbf{u}_0^T \in \mathbb{R}^n$, temperature $\tau$, augmentation sets $\mathcal{P}_1, \mathcal{P}_2$
\STATE Draw a batch of $B$ image-text pairs denoted by $\mathcal{B} = \{(\mathbf{x}_i, \mathbf{z}_i)\}_{i=1}^B$
\FOR{$n = 1 \ldots num\_batches$}
    \FOR{$(\mathbf{x}_i, \mathbf{z}_i) \in \mathcal{B}$}
        \STATE Sample augmentations $\omega_1 \sim \mathcal{P}_1, \omega_2 \sim \mathcal{P}_2$
        \STATE Generate augmented pairs $(\hat{\mathbf{x}}_i, \hat{\mathbf{z}}_i) = (\omega_1(\mathbf{x}_i), \omega_2(\mathbf{z}_i))$
        \STATE Compute image embeddings $E_I(\mathbf{x}_i), E_I(\hat{\mathbf{x}}_i)$
        \STATE Compute text embeddings $E_T(\mathbf{z}_i), E_T(\hat{\mathbf{z}}_i)$
        \STATE Compute cross-modal similarities for all combinations
        \STATE Compute $g(\mathbf{w}_t; \mathbf{x}_i, \mathcal{B}_i)$ and $g(\mathbf{w}_t; \mathbf{z}_i, \mathcal{B}_i)$ for each combination
        \STATE Update $\mathbf{u}_{i,t}^I, \mathbf{u}_{i,t}^T$ according to moving average update rule
    \ENDFOR
    \STATE Compute gradient estimator $\mathbf{m}_t$ across all combinations
    \STATE $\mathbf{v}_t = (1-\beta)\mathbf{v}_{t-1} + \beta\mathbf{m}_t$
    \STATE $\mathbf{w}_{t+1} = \mathbf{w}_t - \eta\mathbf{v}_t$ (using AdamW-style update)
\ENDFOR
\end{algorithmic}
\end{algorithm}

In our study we take $\omega$ = 1, for both AmCLR and xAmCLR experimentations. This results in the eight combinations for contrastive loss for each batch sampled from a dataset. For each of these combinations, we define the contrastive loss in both directions (image-to-text and text-to-image) as expected. Thereby, $\kappa$ = 8 here.
\begin{equation}
F_1(w;B) = -\frac{\tau}{m} \sum_{i=1}^{m} \log \frac{\exp \left( E_I(x_i)^\top E_T(t_i) / \tau \right)}{\sum_{t \neq t_i} \exp \left( E_I(x_i)^\top E_T(t_j) / \tau \right)},
\end{equation}
\begin{equation}
F_2(w;B) = -\frac{\tau}{m} \sum_{i=1}^{m} \log \frac{\exp \left( E_I(x_i)^\top E_T(\hat{t}_i) / \tau \right)}{\sum_{\hat{t} \neq \hat{t}_i} \exp \left( E_I(x_i)^\top E_T(\hat{t}_j) / \tau \right)},
\end{equation}
\begin{equation}
F_3(w;B) = -\frac{\tau}{m} \sum_{i=1}^{m} \log \frac{\exp \left( E_I(\hat{x}_i)^\top E_T(t_i) / \tau \right)}{\sum_{t \neq t_i} \exp \left( E_I(\hat{x}_i)^\top E_T(t_j) / \tau \right)},
\end{equation}
\begin{equation}
F_4(w;B) = -\frac{\tau}{m} \sum_{i=1}^{m} \log \frac{\exp \left( E_I(\hat{x}_i)^\top E_T(\hat{t}_i) / \tau \right)}{\sum_{\hat{t} \neq \hat{t}_i} \exp \left( E_I(\hat{x}_i)^\top E_T(\hat{t}_j) / \tau \right)},
\end{equation}
\begin{equation}
F_5(w;B) = -\frac{\tau}{m} \sum_{i=1}^{m} \log \frac{\exp \left( E_I(x_i)^\top E_T(t_i) / \tau \right)}{\sum_{x \neq x_i} \exp \left( E_I(x_j)^\top E_T(t_i) / \tau \right)},
\end{equation}
\begin{equation}
F_6(w;B) = -\frac{\tau}{m} \sum_{i=1}^{m} \log \frac{\exp \left( E_I(\hat{x}_i)^\top E_T(t_i) / \tau \right)}{\sum_{\hat{x} \neq \hat{x}_i} \exp \left( E_I(\hat{x}_j)^\top E_T(t_i) / \tau \right)},
\end{equation}
\begin{equation}
F_7(w;B) = -\frac{\tau}{m} \sum_{i=1}^{m} \log \frac{\exp \left( E_I(x_i)^\top E_T(\hat{t}_i) / \tau \right)}{\sum_{x \neq x_i} \exp \left( E_I(x_j)^\top E_T(\hat{t}_i) / \tau \right)},
\end{equation}
\begin{equation}
F_8(w;B) = -\frac{\tau}{m} \sum_{i=1}^{m} \log \frac{\exp \left( E_I(\hat{x}_i)^\top E_T(\hat{t}_i) / \tau \right)}{\sum_{\hat{x} \neq \hat{x}_i} \exp \left( E_I(\hat{x}_j)^\top E_T(\hat{t}_i) / \tau \right)}.
\end{equation}
\subsection{xAmCLR: Extension of AmCLR}

In xAmCLR, in addition to AmCLR, we add extra terms for intra-modality learning, where we contrast augmented images with other images and augmented text with other texts. This approach is inspired by the unimodal SogCLR framework, and results in the following $\kappa$ combinations for contrastive loss for each batch sampled from the dataset. The motivation behind this strategy is to enhance representation learning by incorporating unimodal losses alongside the existing multimodal losses. By considering unimodal losses between images and their augmented versions, as well as between text and their paraphrased versions, the model aims to achieve a more generalized representation learning by capturing the nuances within each modality independently.

Thus in total we have now increased the number of combinations as explained below.
\[
\text{Let } \omega \text{ be the number of augmentations per modality.}
\]
\[
\text{So, each modality (image or text) has } (\omega + 1) \text{ variations (original + augmentations).}
\]
\[
\text{Total combinations for both image-to-text and text-to-image are:}
\]
\[
\kappa_{old} = 2 \times (\omega + 1) \times (\omega + 1)
\]
\[
\text{Now, adding combinations for image-to-augmented-image and text-to-augmented-text:}
\]
\[
\kappa = 2 \times (\omega + 1) \times (\omega + 1) + 2 \times \binom{\omega + 1}{2} + 2 \times \binom{\omega + 1}{2}
\]
\[
\text{where } \binom{\omega+1}{2} = \frac{(\omega+1) \cdot \omega}{2}.
\]

New addition to AmCLR's loss is additional intra-modal losses, e.g. \( F_\phi=(w; B) \) represents the loss for intra-image combination, similarly we extend it for all the possible augmentations, both ways, and for texts.

\begin{equation}
F_1(w;B) = -\frac{\tau}{m} \sum_{i=1}^{m} \log \frac{\exp \left( E_I(x_i)^\top E_T(t_i) / \tau \right)}{\sum_{t \neq t_i} \exp \left( E_I(x_i)^\top E_T(t_j) / \tau \right)},
\end{equation}
\begin{equation}
F_2(w;B) = -\frac{\tau}{m} \sum_{i=1}^{m} \log \frac{\exp \left( E_I(x_i)^\top E_T(\hat{t}_i) / \tau \right)}{\sum_{\hat{t} \neq \hat{t}_i} \exp \left( E_I(x_i)^\top E_T(\hat{t}_j) / \tau \right)},
\end{equation}
\begin{equation}
F_3(w;B) = -\frac{\tau}{m} \sum_{i=1}^{m} \log \frac{\exp \left( E_I(\hat{x}_i)^\top E_T(t_i) / \tau \right)}{\sum_{t \neq t_i} \exp \left( E_I(\hat{x}_i)^\top E_T(t_j) / \tau \right)},
\end{equation}
\begin{equation}
F_4(w;B) = -\frac{\tau}{m} \sum_{i=1}^{m} \log \frac{\exp \left( E_I(\hat{x}_i)^\top E_T(\hat{t}_i) / \tau \right)}{\sum_{\hat{t} \neq \hat{t}_i} \exp \left( E_I(\hat{x}_i)^\top E_T(\hat{t}_j) / \tau \right)},
\end{equation}
\begin{equation}
\vdots
\end{equation}
\begin{equation}
F_\phi(w;B) = -\frac{\tau}{m} \sum_{i=1}^{m} \log \frac{\exp \left( E_I(x_i)^\top E_I(\hat{x}_i) / \tau \right)}{\sum_{\hat{x} \neq \hat{x}_i} \exp \left( E_I(x_i)^\top E_I(\hat{x}_j) / \tau \right)},
\end{equation}
\begin{equation}
\vdots
\end{equation}
\begin{equation}
F_\kappa(w; \beta) \text{ (extends similarly...)}.
\end{equation}

The overall dataset loss \( F(w) \) can be expressed as:
\begin{equation}
F(w; B) = \sum_{k=1}^{\kappa} F_k(w; B),
\end{equation}
where \( F_k(w; B) \) corresponds to the $\kappa$ combinations.
The dataset loss \( F(w) \) is:
\begin{equation}
F(w) = \mathbb{E}_{B \sim D} \left[ F(w; B) \right].
\end{equation}
Decomposing \( F(w; B) \):
\begin{equation}
F(w; B) = g_1(w; B) + g_2(w; B),
\end{equation}
where
\begin{equation}
g_1(w; B) = F_p(w; B) + F_q(w; B),
\end{equation}
\begin{equation}
g_2(w; B) = \sum_{k=1, k \neq p, q}^{\kappa} F_k(w; B).
\end{equation}
Here, the notations $p$ and $q$ used are same as for AmCLR. Now the full dataset loss \( F(w) \) can then be expressed as:
\begin{equation}
F(w) = \mathbb{E}_{B \sim \mathcal{D}} \left[ F_p(w; B) + F_q(w; B) \right]
+ \mathbb{E}_{Image \sim \mathcal P1, Text \sim \mathcal P2, B \sim \mathcal{D}} \left[ \sum_{\substack{k=1 \\ k \neq p, q}}^{\kappa} F_k(w; B) \right].
\end{equation}

The gradient estimator \( m_t \) for the full dataset across each batch \( F(w) \) is:
\begin{equation}
m_t = - \frac{1}{|B|} \sum_{(x, t) \in B} E_I(x)^\top E_T(t) + \frac{1}{|B|} \sum_{(x, t) \in B} \frac{\tau}{u_{I_{x,t}}} \nabla g(w_t; x, B) + \frac{\tau}{u_{T_{x,t}}} \nabla g(w_t; t, B),
\end{equation}

where \( g(w; x, B) \) and \( g(w; t, B) \), as seen in algorithm \ref{algo:xAmCLR}, are the accumulated gradients for all the $\kappa$ cases per batch, and the batch estimators of \( g(w; x) \) and \( g(w; t) \), respectively. The scalars \( u_{I_{x,t}} \) and \( u_{T_{x,t}} \) are updated for the sampled data according to:
\begin{equation}
u_{I_{x,t+1}} = (1 - \gamma) u_{I_{x,t}} + \gamma g(w_t; x, B),
\end{equation}
\begin{equation}
u_{T_{x,t+1}} = (1 - \gamma) u_{T_{x,t}} + \gamma g(w_t; t, B),
\end{equation}

Finally, the model parameters are updated using an AdamW-style update rule:
\begin{equation}
w_{t+1} = w_t - \eta v_t,
\end{equation}

where \( v_t \) is the momentum term and \( \eta \) is the learning rate.

\begin{algorithm}[H]
\caption{xAmCLR}
\label{algo:xAmCLR}
\begin{algorithmic}[1]
\REQUIRE $\mathbf{w}_0 \in \mathbb{R}^d$, Initialize $\mathbf{u}_0^I, \mathbf{u}_0^T \in \mathbb{R}^n$, temperature $\tau$, augmentation sets $\mathcal{P}_1, \mathcal{P}_2$
\STATE Draw a batch of $B$ image-text pairs denoted by $\mathcal{B} = \{(\mathbf{x}_i, \mathbf{z}_i)\}_{i=1}^B$
\STATE \textbf{for} $n = 1\ldots num\_batches$ \textbf{do}
    \FOR{$(\mathbf{x}_i, \mathbf{z}_i) \in \mathcal{B}$}
        \STATE Sample augmentations $\omega_1 \sim \mathcal{P}_1, \omega_2 \sim \mathcal{P}_2$
        \STATE Generate augmented pairs $(\hat{\mathbf{x}}_i, \hat{\mathbf{z}}_i) = (\omega_1(\mathbf{x}_i), \omega_2(\mathbf{z}_i))$
        \STATE Compute image embeddings $E_I(\mathbf{x}_i), E_I(\hat{\mathbf{x}}_i)$
        \STATE Compute text embeddings $E_T(\mathbf{z}_i), E_T(\hat{\mathbf{z}}_i)$
        \STATE Compute cross-modal similarities for all combinations
        \STATE Compute intra-modal similarities $\{(\mathbf{x}_i, \hat{\mathbf{x}}_i), (\mathbf{z}_i, \hat{\mathbf{z}}_i)\}$
        \STATE Compute $g(\mathbf{w}_t; \mathbf{x}_i, \mathcal{B}_i)$ and $g(\mathbf{w}_t; \mathbf{z}_i, \mathcal{B}_i)$ for all combinations
        \STATE Update $\mathbf{u}_{i,t}^I, \mathbf{u}_{i,t}^T$ according to moving average update rule
    \ENDFOR
    \STATE Compute gradient estimator $\mathbf{m}_t$ across all combinations 
    \STATE $\mathbf{v}_t = (1-\beta)\mathbf{v}_{t-1} + \beta\mathbf{m}_t$
    \STATE $\mathbf{w}_{t+1} = \mathbf{w}_t - \eta\mathbf{v}_t$ (using AdamW-style update)
\STATE \textbf{end for}
\end{algorithmic}
\end{algorithm}

As mentioned earlier, we take $\omega$ = 1, for xAmCLR experimentations. This results in the twelve combinations for contrastive loss for each batch sampled from a dataset. For each of these combinations, we define the contrastive loss in both directions (image-to-text, text-to-image, image-to-image and text-to-text) as expected. Thereby, $\kappa$ = 12 here.

\begin{equation}
F_1(w;B) = -\frac{\tau}{m} \sum_{i=1}^{m} \log \frac{\exp \left( E_I(x_i)^\top E_T(t_i) / \tau \right)}{\sum_{t \neq t_i} \exp \left( E_I(x_i)^\top E_T(t_j) / \tau \right)},
\end{equation}
\begin{equation}
F_2(w;B) = -\frac{\tau}{m} \sum_{i=1}^{m} \log \frac{\exp \left( E_I(x_i)^\top E_T(\hat{t}_i) / \tau \right)}{\sum_{\hat{t} \neq \hat{t}_i} \exp \left( E_I(x_i)^\top E_T(\hat{t}_j) / \tau \right)},
\end{equation}
\begin{equation}
F_3(w;B) = -\frac{\tau}{m} \sum_{i=1}^{m} \log \frac{\exp \left( E_I(\hat{x}_i)^\top E_T(t_i) / \tau \right)}{\sum_{t \neq t_i} \exp \left( E_I(\hat{x}_i)^\top E_T(t_j) / \tau \right)},
\end{equation}
\begin{equation}
F_4(w;B) = -\frac{\tau}{m} \sum_{i=1}^{m} \log \frac{\exp \left( E_I(\hat{x}_i)^\top E_T(\hat{t}_i) / \tau \right)}{\sum_{\hat{t} \neq \hat{t}_i} \exp \left( E_I(\hat{x}_i)^\top E_T(\hat{t}_j) / \tau \right)},
\end{equation}
\begin{equation}
F_5(w;B) = -\frac{\tau}{m} \sum_{i=1}^{m} \log \frac{\exp \left( E_I(x_i)^\top E_T(t_i) / \tau \right)}{\sum_{x \neq x_i} \exp \left( E_I(x_j)^\top E_T(t_i) / \tau \right)},
\end{equation}
\begin{equation}
F_6(w;B) = -\frac{\tau}{m} \sum_{i=1}^{m} \log \frac{\exp \left( E_I(\hat{x}_i)^\top E_T(t_i) / \tau \right)}{\sum_{\hat{x} \neq \hat{x}_i} \exp \left( E_I(\hat{x}_j)^\top E_T(t_i) / \tau \right)},
\end{equation}
\begin{equation}
F_7(w;B) = -\frac{\tau}{m} \sum_{i=1}^{m} \log \frac{\exp \left( E_I(x_i)^\top E_T(\hat{t}_i) / \tau \right)}{\sum_{x \neq x_i} \exp \left( E_I(x_j)^\top E_T(\hat{t}_i) / \tau \right)},
\end{equation}
\begin{equation}
F_8(w;B) = -\frac{\tau}{m} \sum_{i=1}^{m} \log \frac{\exp \left( E_I(\hat{x}_i)^\top E_T(\hat{t}_i) / \tau \right)}{\sum_{\hat{x} \neq \hat{x}_i} \exp \left( E_I(\hat{x}_j)^\top E_T(\hat{t}_i) / \tau \right)}.
\end{equation}
\begin{equation}
F_9(w;B) = -\frac{\tau}{m} \sum_{i=1}^{m} \log \frac{\exp \left( E_I(x_i)^\top E_I(\hat{x}_i) / \tau \right)}{\sum_{\hat{x} \neq \hat{x}_i} \exp \left( E_I(x_i)^\top E_I(\hat{x}_j) / \tau \right)},
\end{equation}
\begin{equation}
F_{10}(w;B) = -\frac{\tau}{m} \sum_{i=1}^{m} \log \frac{\exp \left( E_I(x_i)^\top E_I(\hat{x}_i) / \tau \right)}{\sum_{x \neq x_i} \exp \left( E_I(x_j)^\top E_I(\hat{x}_i) / \tau \right)},
\end{equation}
\begin{equation}
F_{11}(w;B) = -\frac{\tau}{m} \sum_{i=1}^{m} \log \frac{\exp \left( E_T(t_i)^\top E_T(\hat{t}_i) / \tau \right)}{\sum_{\hat{t} \neq \hat{t}_i} \exp \left( E_T(t_i)^\top E_T(\hat{t}_j) / \tau \right)},
\end{equation}
\begin{equation}
F_{12}(w;B) = -\frac{\tau}{m} \sum_{i=1}^{m} \log \frac{\exp \left( E_T(t_i)^\top E_T(\hat{t}_i) / \tau \right)}{\sum_{t \neq t_i} \exp \left( E_T(t_j)^\top E_T(\hat{t}_i) / \tau \right)},
\end{equation}

\section{Experiments}
\label{others}
In this section, we compare our proposed losses AmCLR and xAmCLR to SogCLR and iSogCLR losses. To ensure fairness, we adopt the same settings for all. We evaluate the performance of different loss functions combined with various optimizers on three tasks: Retrieval (Text), Retrieval (Image), and Zero-shot classification. For each task, we report standard metrics including Top-1, Top-5, and Top-10 accuracy to provide a comprehensive comparison.

\subsection{Setup}
We utilized the NVIDIA RTX 6000 GPU nodes for distributed parallel training and evaluation of our models. The batch size was fixed at 128, with models trained over 30 epochs to ensure convergence. We chose a small batch size to leverage the improvements introduced by SogCLR, which addresses the limitations of traditional contrastive learning models like CLIP that require large batch sizes for effective training. By approximating the global contrastive loss, SogCLR allows us to maintain high accuracy even with smaller batches, thus reducing computational demands and enhancing training efficiency. Our approach aims to further improve upon SogCLR by optimizing performance with this efficient batch size. We experimented with various optimizers including AdamW, AdamP, RAdam, SGDP, NAdam, and NvNovograd to assess their impact on model performance.

\subsection{Datasets}
\textbf{Training data}: We used a 100k subset of the Conceptual Captions 3M dataset for training. This dataset offers a diverse collection of image-text pairs that facilitate effective learning of diverse and discriminative features, making it suitable for contrastive learning.

\textbf{Validation data}: We employed a subset of the MSCOCO dataset to evaluate retrieval performance. MSCOCO's detailed annotations makes it suitable for assessing models' retrieval capabilities. We then employed a subset of the ImageNet dataset to evaluate zero-shot classification performance. ImageNet's extensive collection of annotated images across numerous categories provides a robust benchmark for testing models' ability to classify unseen categories.

\subsection{Model Architecture}

\textbf{Image Encoder}: We utilized a ResNet-50 model pretrained on ImageNet. ResNet-50, comprising of 50 layers and 25.6 million parameters is particularly well-suited for tasks involving complex visual feature extraction, allowing it to classify images into 1,000 different categories, providing a strong foundation for downstream vision-language tasks.

\textbf{Text Encoder}: We utilized DistilBERT, a distilled version of BERT, pretrained on BookCorpus (800M words) and English Wikipedia (2.5B words). DistilBERT, with 66 million parameters, is 40\% smaller and 60\% faster than BERT while retaining 97\% of its language understanding capabilities.

\section{Results}
\begin{figure}[H]
    \centering
    \begin{subfigure}{\textwidth}
        \centering
        \includegraphics[width=1.15\textwidth, height=0.9\textheight, keepaspectratio]{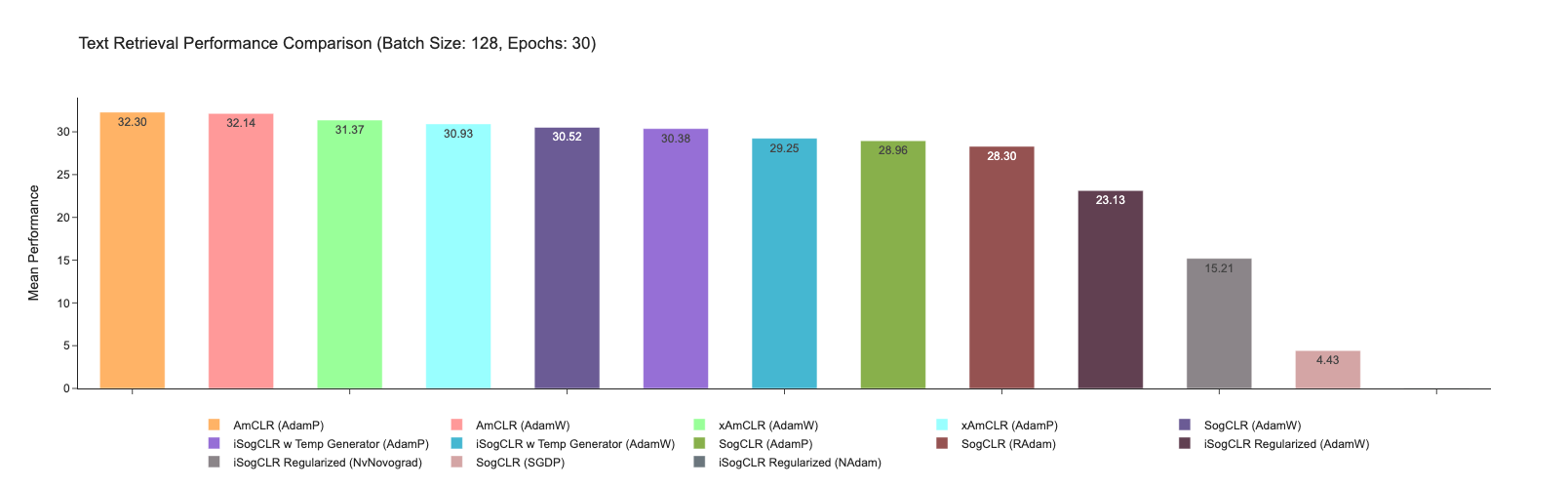}
        \caption{Text Retrieval Performance}
    \end{subfigure}
    \vspace{8mm}
    \begin{subfigure}{\textwidth}
        \centering
        \includegraphics[width=1.15\textwidth, height=0.9\textheight, keepaspectratio]{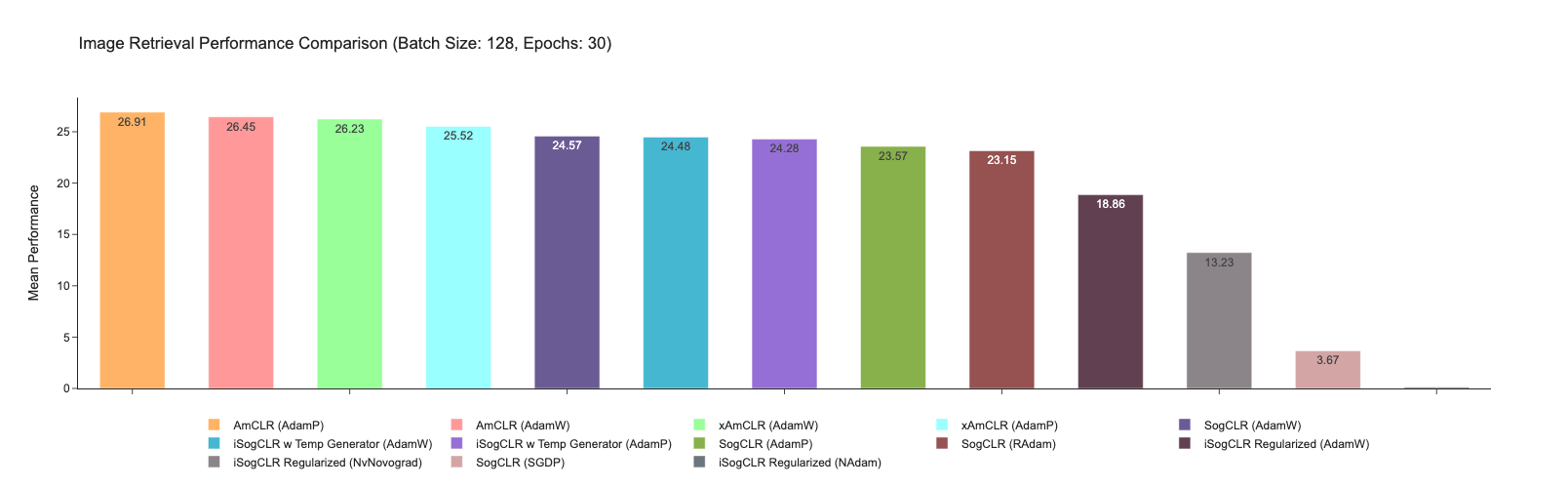}
        \caption{Image Retrieval Performance}
    \end{subfigure}
    \caption{Comprehensive comparison of retrieval performance across text and image modalities. Both AmCLR and xAmCLR consistently outperform baseline methods, with AmCLR (AdamP) achieving highest mean performance of 32.30\% and 26.91\% on text and image retrieval respectively. The results demonstrate the effectiveness of our approaches across different optimizers and modalities while maintaining computational efficiency.}
    \label{fig:retrieval_comparison}
\end{figure}

\begin{figure}[H]
    \centering
    \includegraphics[width=1.15\textwidth, height=0.9\textheight, keepaspectratio]{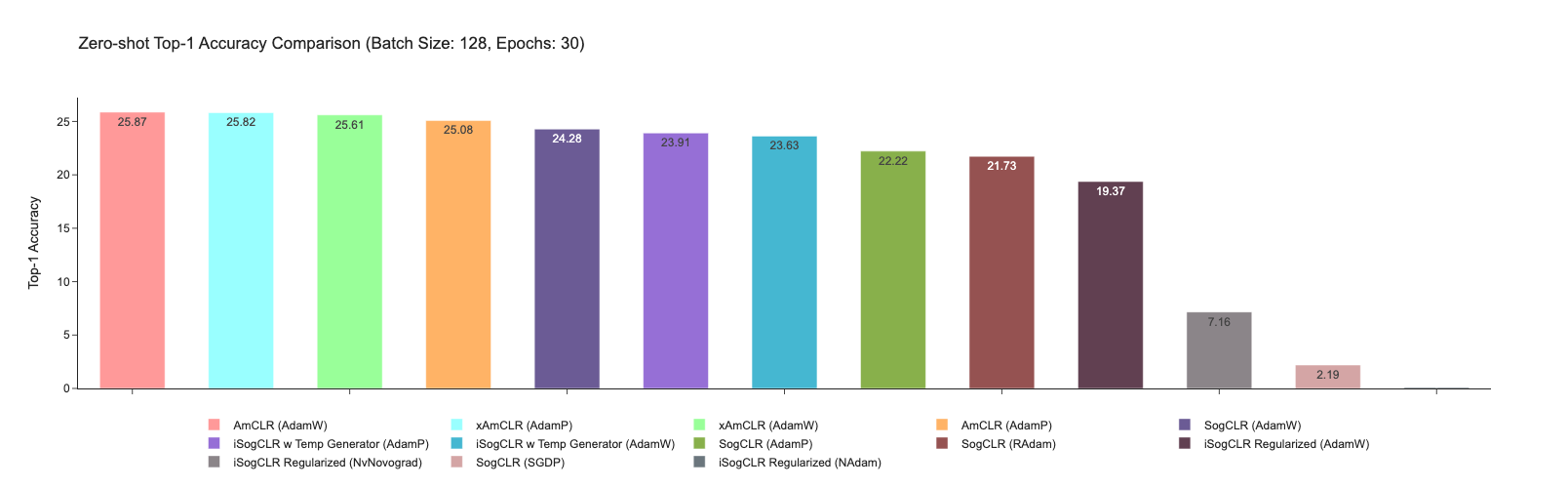}
    \caption{Zero-shot Top-1 Accuracy Comparison. The plot shows the zero-shot learning capabilities of our approaches, with AmCLR (AdamW) achieving 25.87\% accuracy, followed closely by xAmCLR variants. This demonstrates the models' ability to generalize to unseen data without additional training.}
    \label{fig:zeroshot_top1}
\end{figure}

In our preliminary experiments, we observed that optimizers such as RAdam, NAdam, NvNovograd and SGDP performed poorly when used with existing loss functions like SogCLR and iSogCLR. This suboptimal performance can be attributed to their inability to effectively handle the stochasticity of smaller batch sizes. Consequently, we chose not to experiment with these optimizers for our new loss functions, AmCLR and xAmCLR, focusing instead on AdamW and AdamP, which showed more promising results in initial tests. We have used two versions of iSogCLR loss in our experiments, one which uses a temperature generator to dynamically adjust the temperatures for image and text features and the other which introduces a regularization term to control the loss from positive pairs. Figure's \ref{fig:retrieval_comparison} and \ref{fig:zeroshot_top1} show the comparison.

\subsection{Retrieval (Text) Tasks}

Table \ref{tab:retrieval_text} presents a detailed comparison of performance metrics for text retrieval tasks. Our proposed AmCLR with AdamW achieved a Top-1 accuracy of 14.64\%, significantly outperforming SogCLR with the same optimizer, which achieved 13.1\%. Similarly, AmCLR with AdamP reached a Top-1 accuracy of 14.54\%, surpassing iSogCLR\_New's best performance of 13.14\% with AdamP. In terms of broader retrieval metrics, AmCLR consistently led in Top-5 and Top-10 categories, confirming its superior generalization capabilities.

xAmCLR also demonstrated strong performance with both optimizers, achieving a Top-1 accuracy of 14.14\% with AdamW and 13.62\% with AdamP, maintaining a competitive edge over other losses.

\subsection{Retrieval (Image) Tasks}

As shown in Table \ref{tab:retrieval_image}, the image retrieval tasks further highlight the advantages of our solutions. AmCLR with AdamP achieved the highest Top-1 accuracy at 11.46\%, compared to iSogCLR\_New's 10.13\% with AdamP. This trend continued across Top-5 and Top-10 metrics, where AmCLR consistently outperformed other losses by substantial margins. Additionally, we observed that AdamP performed slightly better than AdamW across image retrieval tasks.

xAmCLR closely followed AmCLR, achieving Top-1 accuracy at 11.14\% with AdamW. Similar trends were observed for Top-5 and Top-10 metrics, with xAmCLR results within 0.3–0.7 percentage points of AmCLR.

\subsection{Zero-shot Tasks}

The zero-shot classification tasks presented in Table \ref{tab:zero_shot} reveal that AmCLR with AdamW reached a Top-1 accuracy of 25.87\%, outperforming all other configurations including SogCLR's 24.28\% with AdamW. The results were consistent across Top-5 and Top-10 metrics, where AmCLR demonstrated superior adaptability to unseen data distributions.

xAmCLR also performed exceptionally well in zero-shot tasks, with results nearly matching those of AmCLR, highlighting its effectiveness in scenarios requiring high generalization.\\

Overall, our proposed loss functions, AmCLR and xAmCLR, consistently outperformed existing methods across all tasks and metrics when paired with both AdamW and AdamP optimizers. This underscores their potential as robust solutions for diverse retrieval and classification challenges.

\begin{table}
\centering
\caption{Performance comparison of different losses and optimizers for Retrieval (Text) Tasks.}
\label{tab:retrieval_text}
\setlength{\tabcolsep}{8pt} 
\renewcommand{\arraystretch}{1.1} 
\footnotesize 
\begin{tabular}{llcccc}
\toprule
\textbf{Method} & \textbf{Optimizer} & \textbf{Top-1} & \textbf{Top-5} & \textbf{Top-10} & \textbf{Mean} \\ 
\midrule
SogCLR        & AdamW              & 13.1           & 33.36          & 45.1        & 30.52    \\
SogCLR        & AdamP              & 12              & 31.68              & 43.2     & 28.96          \\
SogCLR        & RAdam              & 11.82          & 30.82          & 42.26      & 28.3     \\
SogCLR        & SGDP               & 1              & 4.5            & 7.78       & 4.43     \\
iSogCLR w Temp Generator  & AdamW              & 12.36          & 32.18          & 43.22      & 29.25     \\
iSogCLR w Temp Generator  & AdamP              & 13.14          & 33.14          & 44.86       & 30.38    \\
iSogCLR Regularized & AdamW           & 8.78           & 24.58          & 36.04        & 23.13   \\
iSogCLR Regularized & NAdam           & 0.02           & 0.04           & 0.06        & 0.04    \\
iSogCLR Regularized & NvNovograd      & 5.18           & 16.54          & 25.14        & 15.21   \\
\midrule
\textbf{AmCLR (ours)}         & \textbf{AdamW}              & \textbf{14.64}          & \textbf{35}             & \textbf{46.78}      & \textbf{32.14}     \\
\textbf{AmCLR (ours)}         & \textbf{AdamP}              & \textbf{14.54}              & \textbf{35.34}               & \textbf{47}        & \textbf{32.30}       \\
\textbf{xAmCLR (ours)}        & \textbf{AdamW}              & \textbf{14.14}          & \textbf{34.24}          & \textbf{45.74}      & \textbf{31.37}     \\
\textbf{xAmCLR (ours)}        & \textbf{AdamP}              & \textbf{13.62}              & \textbf{33.4}               & \textbf{45.78}    & \textbf{30.93}           \\
\bottomrule
\end{tabular}
\end{table}

\begin{table}
\centering
\caption{Performance comparison of different losses and optimizers for Retrieval (Image) Tasks.}
\label{tab:retrieval_image}
\setlength{\tabcolsep}{8pt} 
\renewcommand{\arraystretch}{1.1} 
\footnotesize 
\begin{tabular}{llcccc}
\toprule
\textbf{Method} & \textbf{Optimizer} & \textbf{Top-1} & \textbf{Top-5} & \textbf{Top-10} & \textbf{Mean} \\ 
\midrule
SogCLR        & AdamW              & 10.06          & 26.3           & 37.34       & 24.57    \\
SogCLR        & AdamP              & 9.32              & 25.47              & 35.93    & 23.57           \\
SogCLR        & RAdam              & 9.17           & 24.95          & 35.34      & 23.15     \\
SogCLR        & SGDP               & 0.85           & 3.71           & 6.43         & 3.67   \\
iSogCLR w Temp Generator  & AdamW              & 9.9            & 26.47          & 37.05       & 24.48    \\
iSogCLR w Temp Generator  & AdamP              & 10.13          & 26.1           & 36.62      & 24.28     \\
iSogCLR Regularized & AdamW           & 6.9            & 20.03          & 29.65      & 18.86     \\
iSogCLR Regularized & NAdam           & 0.02           & 0.1            & 0.2         & 0.11    \\
iSogCLR Regularized & NvNovograd      & 4.24           & 13.81          & 21.64       & 13.23    \\
\midrule
\textbf{AmCLR (ours)}         & \textbf{AdamW}              & \textbf{11.08}          & \textbf{28.63}          & \textbf{39.64}         & \textbf{26.45}  \\
\textbf{AmCLR (ours)}         & \textbf{AdamP}              & \textbf{11.46}              & \textbf{29.19}               & \textbf{40.09}     & \textbf{26.91}          \\
\textbf{xAmCLR (ours)}        & \textbf{AdamW}              & \textbf{11.14}          & \textbf{28.21}          & \textbf{39.33}       & \textbf{26.23}    \\
\textbf{xAmCLR (ours)}        & \textbf{AdamP}              & \textbf{10.43}              & \textbf{27.45}               & \textbf{38.69}     & \textbf{25.52}          \\
\bottomrule
\end{tabular}
\end{table}

\begin{table}[H]
\centering
\caption{Performance comparison of different losses and optimizers for Zero-shot tasks.}
\label{tab:zero_shot}
\setlength{\tabcolsep}{8pt} 
\renewcommand{\arraystretch}{1.1} 
\footnotesize 
\begin{tabular}{llccc}
\toprule
\textbf{Method} & \textbf{Optimizer} & \textbf{Top-1} & \textbf{Top-5} & \textbf{Top-10} \\ 
\midrule
SogCLR        & AdamW              & 24.28          & 42.93          & 50.29           \\
SogCLR        & AdamP              & 22.22              & 40.78              & 48.34               \\
SogCLR        & RAdam              & 21.73          & 40.03          & 47.53           \\
SogCLR        & SGDP               & 2.19           & 7.64           & 12.06           \\
iSogCLR w Temp Generator  & AdamW              & 23.63          & 42.26          & 49.64           \\
iSogCLR w Temp Generator  & AdamP              & 23.91          & 42.19          & 49.46           \\
iSogCLR Regularized & AdamW           & 19.37          & 39.25          & 47.33           \\
iSogCLR Regularized & NAdam           & 0.1            & 0.5            & 1               \\
iSogCLR Regularized & NvNovograd      & 7.16           & 20             & 28.11           \\
\midrule
\textbf{AmCLR (ours)}         & \textbf{AdamW}              & \textbf{25.87}          & \textbf{44.34}          & \textbf{50.89}           \\
\textbf{AmCLR (ours)}         & \textbf{AdamP}              & \textbf{25.08}              & \textbf{43.56}               & \textbf{50.42}               \\
\textbf{xAmCLR (ours)}        & \textbf{AdamW}              & \textbf{25.61}          & \textbf{44.07}          & \textbf{50.85}           \\
\textbf{xAmCLR (ours)}        & \textbf{AdamP}              & \textbf{25.82}              & \textbf{44.06}               & \textbf{50.87}               \\
\bottomrule
\end{tabular}
\end{table}

\section{Conclusion}
\label{sec:conclusion}

In this work, we analyzed the performance of various optimizers combined with existing and proposed loss functions for Retrieval (Text), Retrieval (Image), and Zero-shot Classification tasks. Our experiments revealed that optimizers such as RAdam, NAdam, NvNovograd, and SGDP exhibited suboptimal performance when paired with existing loss functions like SogCLR and iSogCLR. Based on these findings, we limited our experiments with the proposed loss functions, AmCLR and xAmCLR, to AdamW and AdamP optimizers.

Both AmCLR and xAmCLR consistently outperformed the existing loss functions across all evaluation metrics, in all tasks. While AdamW showed marginally better performance in Zero-shot Classification tasks, AdamP had a slight advantage in Retrieval (Text) and Retrieval (Image) tasks.

Among the proposed loss functions, AmCLR generally achieved higher accuracies compared to xAmCLR. However, the performance gap between the two was notably smaller in Zero-shot Classification tasks, suggesting that xAmCLR excels at generalization. These results highlight the potential of AmCLR and xAmCLR as robust loss functions for bimodal contrastive learning.

\section{Future Research}

\begin{itemize}
    \item Our initial experiments with AmCLR and xAmCLR demonstrate promising results on a 100K subset of the Conceptual Captions 3M dataset. The significant performance improvements achieved with relatively constrained computational resources—a batch size of 128 and 30 epochs—strongly suggest that scaling up these approaches could yield substantially better results. We plan to extend our research in several key directions.
    
    \item First, we anticipate that training on the complete CC3M dataset of 3 million image-text pairs will enable our models to learn more nuanced and robust representations. The strong performance on the current subset indicates that our architectural choices and loss functions can effectively capture cross-modal relationships. With access to the full dataset, we expect to achieve more comprehensive coverage of semantic relationships and improved generalization capabilities. This expansion will require careful hyper-parameter optimization, particularly for batch sizes and epoch counts, to fully leverage the increased data volume while maintaining computational efficiency.
    
    \item A critical focus of our future work will be the development of a robust distributed training infrastructure. We envision implementing a hierarchical distributed training architecture with primary and secondary coordinator nodes managing multiple worker clusters. This system will incorporate sophisticated fault tolerance mechanisms, including checkpoint synchronization protocols and automated worker node recovery. We plan to implement a distributed logging system that aggregates training metrics, system health indicators, and resource utilization data in real-time. To ensure training stability, we will deploy an automated monitoring system that tracks gradient norms, loss convergence patterns, and cross-node consistency metrics. The infrastructure will include fallback mechanisms such as gradient accumulation buffers and dynamic batch size adjustment to handle temporary node failures without compromising training integrity. Additionally, we will implement distributed data loading pipelines with pre-fetching and caching mechanisms to optimize I/O operations across the training cluster.
    
    \item Our current implementation uses a conservative augmentation strategy with $\omega = 1$. Future research will explore the impact of increasing the number and diversity of augmentations. For images, we plan to investigate more sophisticated transformation pipelines that preserve semantic content while creating challenging positive pairs. This includes implementing adaptive augmentation strategies that adjust transformation intensity based on training dynamics. For text modality, we aim to use more nuanced paraphrasing techniques that generate semantically equivalent but syntactically diverse expressions. These enhanced augmentation strategies should create more challenging negative samples while maintaining semantic consistency, thereby improving the model's ability to learn robust cross-modal representations.
    
    \item Finally, we recognize the potential synergy between our augmentation-based approaches and the distributionally robust optimization (DRO) framework employed in iSogCLR. We hypothesize that combining these approaches could yield a more robust training objective that benefits from both individual temperature optimization and diverse augmentation strategies. We plan to extend our framework to incorporate DRO principles, potentially leading to a unified approach that leverages the strengths of both methodologies. This integration could potentially address the challenges of varying semantic granularity in cross-modal learning while maintaining computational efficiency.
\end{itemize}



\medskip

{
\small
\begin{thebibliography}{99}
  \bibitem{chen2020simple}
    Chen, T., Kornblith, S., Norouzi, M., \& Hinton, G. (2020). A simple framework for contrastive learning of visual representations. In \textit{International Conference on Machine Learning}, pp. 1597--1607.
    
  \bibitem{radford2021learning}
    Radford, A., Kim, J. W., Hallacy, C., Ramesh, A., Goh, G., Agarwal, S., Sastry, G., Askell, A., Mishkin, P., Clark, J., \& others. (2021). Learning transferable visual models from natural language supervision. In \textit{International Conference on Machine Learning}, pp. 8748--8763.

  


  \bibitem{deng2009imagenet}
    Deng, J., Dong, W., Socher, R., Li, L.-J., Li, K., \& Fei-Fei, L. (2009). ImageNet: A large-scale hierarchical image database. In \textit{2009 IEEE Conference on Computer Vision and Pattern Recognition}, pp. 248--255.

  \bibitem{lin2014microsoft}
    Lin, T.-Y., Maire, M., Belongie, S., Hays, J., Perona, P., Ramanan, D., Doll{\'a}r, P., \& Zitnick, C. L. (2014). Microsoft COCO: Common objects in context. In \textit{Computer Vision--ECCV 2014: 13th European Conference}, pp. 740--755.

    
  \bibitem{yuan2022provable}
    Yuan, Z., Wu, Y., Qiu, Z.-H., Du, X., Zhang, L., Zhou, D., \& Yang, T. (2022). Provable stochastic optimization for global contrastive learning: Small batch does not harm performance. In \textit{International Conference on Machine Learning}, pp. 25760--25782.

  \bibitem{qiu2023not}
    Qiu, Z.-H., Hu, Q., Yuan, Z., Zhou, D., Zhang, L., \& Yang, T. (2023). Not all semantics are created equal: Contrastive self-supervised learning with automatic temperature individualization. \textit{arXiv preprint arXiv:2305.11965}.

    \bibitem{mehta2023benchmarking}
    Mehta, A., Sengupta, P., Garg, D., Singh, H., \& Diamand, Y. S. (2023). Benchmarking the Effectiveness of Classification Algorithms and SVM Kernels for Dry Beans. \textit{arXiv preprint arXiv:2307.07863}.
    

  \bibitem{sharma2018conceptual}
    Sharma, P., Ding, N., Goodman, S., \& Soricut, R. (2018). Conceptual captions: A cleaned, hypernymed, image alt-text dataset for automatic image captioning. In \textit{Proceedings of the 56th Annual Meeting of the Association for Computational Linguistics}, pp. 2556--2565.

  \bibitem{he2016deep}
    He, K., Zhang, X., Ren, S., \& Sun, J. (2016). Deep residual learning for image recognition. In \textit{Proceedings of the IEEE Conference on Computer Vision and Pattern Recognition}, pp. 770--778.

  \bibitem{sanh2019distilbert}
    Sanh, V. (2019). DistilBERT, a distilled version of BERT: Smaller, faster, cheaper, and lighter. \textit{arXiv preprint arXiv:1910.01108}.

  \bibitem{loshchilov2017decoupled}
    Loshchilov, I. (2017). Decoupled weight decay regularization. \textit{arXiv preprint arXiv:1711.05101}.

    \bibitem{heo2020adamp}
    Heo, B., Chun, S., Oh, S. J., Han, D., Yun, S., Kim, G., Uh, Y., \& Ha, J.-W. (2020). Adamp: Slowing down the slowdown for momentum optimizers on scale-invariant weights. \textit{arXiv preprint arXiv:2006.08217}.

  
  

  \bibitem{gan2022vision}
    Gan, Z., Li, L., Li, C., Wang, L., Liu, Z., Gao, J., \& others. (2022). Vision-language pre-training: Basics, recent advances, and future trends. \textit{Foundations and Trends{\textregistered} in Computer Graphics and Vision}, 14(3--4), 163--352.

  
  \bibitem{mehta2023multi}
    Mehta, A., Sengupta, P., \& Rana, P. S. (2023). A Multi-layered Approach to Brain Tumor Classification Using VDC-12. In \textit{International Conference on Computational Sciences and Sustainable Technologies}, pp. 379--391.

  \bibitem{goel2022cyclip}
    Goel, S., Bansal, H., Bhatia, S., Rossi, R., Vinay, V., \& Grover, A. (2022). Cyclip: Cyclic contrastive language-image pretraining. \textit{Advances in Neural Information Processing Systems}, 35, 6704--6719.

  \bibitem{zhang2022align}
    Zhang, S., Qiu, L., Zhu, F., Yan, J., Zhang, H., Zhao, R., Li, H., \& Yang, X. (2022). Align representations with base: A new approach to self-supervised learning. In \textit{Proceedings of the IEEE/CVF Conference on Computer Vision and Pattern Recognition}, pp. 16600--16609.

  \bibitem{smeu2024declip}
    Smeu, S., Oneata, E., \& Oneata, D. (2024). DeCLIP: Decoding CLIP representations for deepfake localization. \textit{arXiv preprint arXiv:2409.08849}.

  \bibitem{mu2022slip}
    Mu, N., Kirillov, A., Wagner, D., \& Xie, S. (2022). Slip: Self-supervision meets language-image pre-training. In \textit{European Conference on Computer Vision}, pp. 529--544.

  \bibitem{yao2021filip}
    Yao, L., Huang, R., Hou, L., Lu, G., Niu, M., Xu, H., Liang, X., Li, Z., Jiang, X., \& Xu, C. (2021). Filip: Fine-grained interactive language-image pre-training. \textit{arXiv preprint arXiv:2111.07783}.

  \bibitem{rusak2024infonce}
    Rusak, E., Reizinger, P., Juhos, A., Bringmann, O., Zimmermann, R. S., \& Brendel, W. (2024). InfoNCE: Identifying the Gap Between Theory and Practice. In \textit{High-dimensional Learning Dynamics 2024: The Emergence of Structure and Reasoning}.

 


  
\end{thebibliography}
}

\end{document}